\documentclass[journal]{IEEEtran}
\usepackage{graphicx}
\usepackage{amsmath,amssymb} 
\usepackage{color}
\usepackage{multirow}

\begin{document}
%
\title{Deep Aesthetic Quality Assessment with Semantic Information}
%
%

\author{
Yueying~Kao,
Ran~He,
Kaiqi~Huang
\thanks{Yueying~Kao, Ran~He and Kaiqi~Huang are with the National Laboratory of Pattern Recognition, Center for Research on Intelligent Perception and Computing, Institute of Automation, Chinese Academy of Sciences, Beijing 100190, China, and also with the University of Chinese Academy of Sciences, Beijing, 100049, China. Ran~He and Kaiqi~Huang are also with the CAS Center for Excellence in Brain Science and Intelligence Technology, Beijing 100190, China (e-mail: yueying.kao@nlpr.ia.ac.cn; rhe@nlpr.ia.ac.cn; kqhuang@nlpr.ia.ac.cn.}
}

\maketitle

\begin{abstract}
  Human beings often assess the aesthetic quality of an image coupled with the identification of the image's semantic content. This paper addresses the correlation issue between automatic aesthetic quality assessment and semantic recognition. We cast the assessment problem as the main task among a multi-task deep model, and argue that semantic recognition task offers the key to address this problem. Based on convolutional neural networks, we employ a single and simple multi-task framework to efficiently utilize the supervision of aesthetic and semantic labels. A correlation item between these two tasks is further introduced to the framework by incorporating the inter-task relationship learning. This item not only provides some useful insight about the correlation but also improves assessment accuracy of the aesthetic task. Particularly, an effective strategy is developed to keep a balance between the two tasks, which facilitates to optimize the parameters of the framework. Extensive experiments on the challenging AVA dataset and Photo.net dataset validate the importance of semantic recognition in aesthetic quality assessment, and demonstrate that multi-task deep models can discover an effective aesthetic representation to achieve state-of-the-art results.
\end{abstract}

\begin{IEEEkeywords}
Visual aesthetic quality assessment, semantic information, multi-task learning.
\end{IEEEkeywords}

%
\IEEEpeerreviewmaketitle

\section{Introduction \label{sec:intro}}
Aesthetic image analysis has attracted increasing attention in computer vision community~\cite{Datta08,Joshi13,Tang13,Marchesotti2015,Siahaan2016,Segalin2016,tarvainen2014content,park2015consensus}. It is related to the high-level perception of visual aesthetics. Machine learning models for visual aesthetic quality assessment have shown to be useful in many applications, e.g., image retrieval, photo management, image editing, and photography~\cite{Datta07,Ke06,Hong2016,zhang2014actively}. Since visual aesthetics is a subjective attribute, automatically assessing aesthetic quality of images is still challenging. Many data-driven approaches~\cite{Datta06,Luo08,Dhar11,Mar11,Yeh13,Tang13,wang13,zhang2014,wu11,zhang2014perception} have been proposed to address this issue. These methods often learn from the aesthetic quality of images that are labeled by humans. Most of these methods aim to discover a meaningful and better aesthetic representation, and often formulate the representation learning as a single and standalone classification task.

Handcrafted features are earlier attempts. They are based on the intuitions of how people perceive the aesthetic quality of images or photographic rules. These features include color~\cite{Ke06,Datta06,Nishiyama2011}, the rule of thirds~\cite{Datta06}, simplicity~\cite{Luo08,Tang13}, and composition~\cite{Dhar11}. Later, generic image descriptors such as bag-of-visual-words (BOV)~\cite{csurka2004visual} and fisher vectors (FV)~\cite{jaakkola1999exploiting} are used to assess aesthetic quality. They are shown to outperform the traditional handcrafted features~\cite{Mar11,Murray12,marchesotti2013learning}. Recently, deep convolutional neural networks (CNNs)~\cite{Krizhevsky12,Zeiler2014} have been applied to aesthetic quality assessment~\cite{lu14,kao,lu2015deep,mai2016composition}. Nevertheless, these computational approaches provide either accurate or interpretable results~\cite{Marchesotti2015}.
\begin{figure*}
  \centering
  \includegraphics[width=18cm]{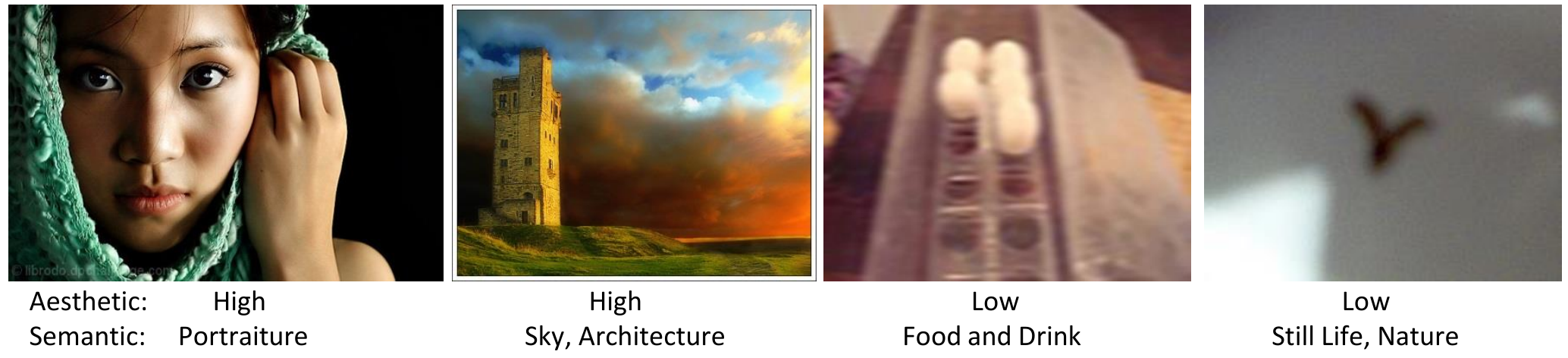}\\
  \caption{Example images with their aesthetic and semantic labels on AVA dataset.
  \label{fig:DAR1}}
\end{figure*}

For human beings, aesthetic quality assessment is always coupled with the identification of semantic content of images~\cite{mullin2015there,locher2015aesthetic}. It is difficult for humans to treat aesthetic quality assessment as an isolate and independent task. When humans assess the aesthetic quality of an image, they first understand what they are assessing. That is, they have known the sematic information of this image. Seen from Fig.~\ref{fig:DAR1}, we can recognize the semantic content from these images at a glance and assess the aesthetic quality quickly. Hence it is reasonable to assume that, assessing aesthetic quality and semantic recognition are correlated tasks for machine learning. However, the relationship between semantic recognition and automatically assessing visual aesthetic quality has not been fully explored.

This paper addresses the correlation issue between automatic aesthetic quality assessment and semantic recognition. We employ multi-task convolutional neural network to explore the potential correlation. Multi-task learning can learn multiple related tasks in parallel with shared knowledge. It has been demonstrated that this approach can boost some or all of the tasks~\cite{caruana1997}. Our goal is to utilize semantic recognition in the joint objective function to improve the aesthetic quality assessment, our main task. However, there is still a typical challenge in the multi-task learning for our multi-task problem. That is, the aesthetic task and semantic task face the different learning difficulties. The main reason is that the semantic recognition is much easier than aesthetics assessment. The semantic content is much objective, while the aesthetic attributes are subjective. Thus, different from the strategies of treating all tasks equally and early stopping~\cite{caruana1997,jung2015rotating,zhang2014facial} we present a strategy to keep the effect of both tasks balanced in the joint objective function.

In addition, to discover the relationships between aesthetic and semantic tasks automatically and to better exploit the inter-task relatedness for more effective feature learning, we model the task relationship and impose it in the objective function. To some extent, it can explain the factors in aesthetic quality assessment and make our results more interpretable. Thus, to investigate how to make full use of semantic information and how semantic information influence aesthetic task, our multi-task framework considers the strategy of keeping the effect of two tasks balanced and the relationship learning between semantic and aesthetic tasks.

In the evaluation, the most challenging large-scale AVA dataset~\cite{Murray12} is used to verify the effectiveness of semantic information for aesthetic feature learning and investigate the correlation among aesthetic and semantic content recognitions. The experiments show that our results significantly outperform the state-of-the-art results~\cite{lu14,lu2015deep,mai2016composition} for aesthetic quality assessment on AVA dataset. Furthermore, it is demonstrated that the learned representation with our multi-task framework can be transferred for the dataset (here we use Photo.net dataset~\cite{Datta08,Datta06}) with only aesthetic labels and other semantic representation (such as from Imagenet) can also be used for aesthetic representation learning.

Our contributions lie in three-fold:
\begin{itemize}
    \item Instead of taking visual aesthetic quality assessment as an isolated task, we propose to exploit the semantic recognition to jointly assess the aesthetic quality with a single multi-task convolutional neural network (MTCNN). It is a novel attempt to learn aesthetic features with the help of a related task, i.e. semantic recognition.
    \item We propose to automatically learn the correlations between the aesthetic and semantic tasks by simultaneously modeling the inter-task relationship and controlling the parameters' complexity of each task in our multi-task framework. It can explain the factors in aesthetic quality assessment and makes our results more interpretable.
    \item Facing the different learning difficulties between the two tasks, we present a strategy to keep the effect of both tasks balanced in the joint objective function. The proposed method outperforms the state-of-the-art methods on the challenging AVA dataset and Photo.net dataset.
\end{itemize}

The rest of this paper is organized as follows: we summarize related work in Section~\ref{sec:rework}, describe our method in detail in Section~\ref{sec:method}, present the experiments in Section~\ref{sec:exp}, and conclude the paper in Section~\ref{sec:con}.

\section{Related work \label{sec:rework}}
Since our work is related to the aesthetic quality assessment and multi-task learning, we will mainly review work related to the two parts in this section.

\subsection{Aesthetic quality assessment}

Most previous works~\cite{Datta06,Ke06,Dhar11,Mar11,Luo11,Niu12} on aesthetic quality assessment focus on the challenging problem of designing appropriate features. Typically, handcrafted features are proposed based on the intuitions about human perception of the aesthetic quality of images or photographic rules. For example, Datta et al.~\cite{Datta06} design certain visual features such as colorfulness, the rule of thirds, and low depth of field indicators, to discriminate between aesthetically pleasing and displeasing images. Dhar et al.~\cite{Dhar11} extract some high level attributes including compositional, content, and sky-illumination attributes, which are characteristically used by humans to describe images. Luo et al.~\cite{Luo11} and Tang et al.~\cite{Tang13} consider that photos may have different aesthetic criteria in mind for different type of images and design visual features in different ways according to the variety of photo content. In~\cite{Mar11}, generic image descriptors are used to assess aesthetic quality, which are shown to outperform the traditional handcrafted features. 

Despite the success of handcrafted features and generic image descriptors, CNNs have been applied to aesthetic quality assessment~\cite{lu14,kao,lu2015deep,mai2016composition} and obtain the state-of-the-art performance. CNNs learn aesthetic features automatically. However, they extract features by treating aesthetic quality assessment as an independent problem. The network in~\cite{lu14}, RDCNN, hopes to leverage the idea of multi-task learning with the style attributes to help determine the aesthetic quality of images. Unfortunately, due to many missing labels for style attributes, they can not jointly perform aesthetics categorization and style classification in a neural network, and just concatenate the features of the aesthetics and style by using transfer learning. Our work is also related to CNNs for aesthetics classification. In contrast, firstly, we exploit semantic information to assist in learning aesthetic representation with a multi-task learning framework. We can jointly learn aesthetics categorization and semantic recognition with a single multi-task network, which is different from RDCNN~\cite{lu14}. Secondly, our multi-task CNN considers the strategy of keeping the effect of two tasks balanced and the relationship learning between semantic and aesthetic tasks. Finally, images are labeled with semantic information much easier than style attributes in real world. This is because only professional photographer and photography amateurs are familiar with all the style attributes.

\begin{figure*}
  \centering
  \includegraphics[width=17cm]{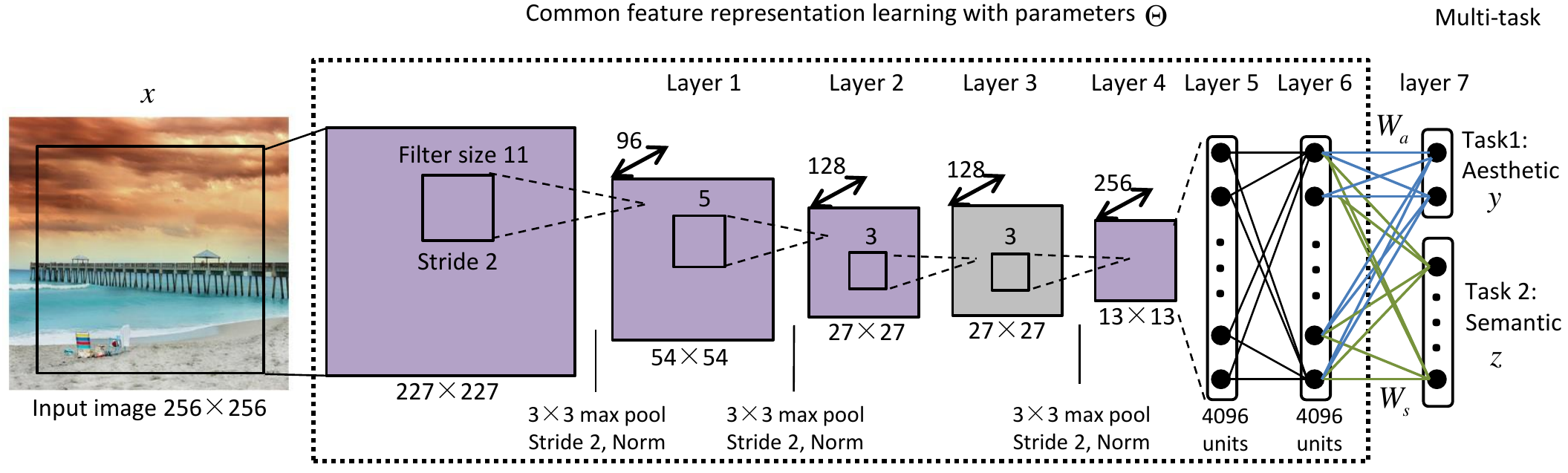}\\
  \caption{An illustration for the architecture of our MTCNN \#1.}
  \label{fig:MT1}
\end{figure*}
\subsection{Multi-task learning}
Multi-task learning aims to boost the generalization performance by learning multiple related tasks simultaneously~\cite{caruana1997,liu2015multi,zhang2014facial,abdulnabi2015multi}. It does this by learning tasks in parallel while using a shared representation~\cite{caruana1997}. Deep neural network can learn features jointly under multiple objectives and it is the earliest models for multi-task learning. Multi-task learning based on deep neural network has been applied to many computer vision problems~\cite{zhang2014facial,jung2015rotating,zhangshu}. However, there are many strategies for sharing knowledge and learning process for different problems. For example, Zhang et al. \cite{zhang2015deep} share parameters in all layers and learn the common features for all tasks, while Liu et al. \cite{export:241128} just sharing in some bottom layers and learn respective representation in some top layers for each task. Yim et al.~\cite{jung2015rotating} treat all tasks equally important. In contrast, early stopping strategy is used in some related tasks~\cite{zhang2014facial}, due to different learning difficulties and convergence rates in different tasks. In our problem, because semantic recognition task is much easier than aesthetic quality assessment, common features of our two tasks are learned simultaneously and an effective strategy of keeping effect of all the tasks balanced in the joint objective function is used. In addition, the task relationships can be learned from the data automatically in the conventional methods~\cite{zhang2010convex,saha2011online,bonilla2007multi}. Inspired by this, we consider the relationship learning in our multi-task neural networks to explore the relationships between the aesthetic and semantic tasks.


\section{Method \label{sec:method}}
In this section, we propose to exploit the semantic information to help identify the aesthetic quality of images, assuming that they are considered as the related attributes~\cite{mullin2015there,locher2015aesthetic}. Here the aesthetic quality assessment is our main task and the semantic content recognition is the aided task. Our problem is firstly formulated as a multi-task convolutional neural network (MTCNN) model without learning task relationships automatically from data. Then we develop a multi-task relationship learning convolutional neural network (MTRLCNN) model by adding the task relationship learning in the objective function to discover the correlation between aesthetic task and semantic tasks. An example of MTCNN architectures is illustrated in Fig.~\ref{fig:MT1}. Furthermore, we explore and adapt different network structures to our problem.

\subsection{Multi-Task Probabilistic Framework \label{sec:MTPF}}
Our problem can be interpreted as a probabilistic model. Using the probabilistic formulation, various deep networks can solve our problem by optimizing the model parameters that maximize the posterior probability. Then, Bayesian analysis is leveraged to predict most likely aesthetic quality and semantic attributes of given images.

Assuming a training dataset with a total of $N$ samples, which are associated with $C$ aesthetic classes and $M$ semantic attributes. Considering each image has only one aesthetic class and multiple semantic attributes in real world, each image is represented as $(x_n, y_n, z_n), n=1,2,...,N$. Here $x_n$ represents the $n$-th image sample, $y_n=c, c=0,..., C-1$ is the aesthetic label and $z_n=[z_n^1,...,z_n^m,...,z_n^M]^{\mathrm{T}}$ is the semantic label for the $n$-th image sample. If the $n$-th image sample has the $m$-th semantic attribute, the $m$-th semantic label is set as $z_n^m=1$, otherwise $z_n^m=0$. Therefore a given dataset is denoted as $(X, Y, Z)=\{(x_n, y_n, z_n), n\in\{1,2,...,N\}\}$. For our MTCNNs (our MTCNN \#1 is shown in Fig.~\ref{fig:MT1}), $\Theta$ denotes the common parameters in some bottom layers to learn features for all tasks, and $W=[W_a, W_s]$ indicates the specific parameters for associated tasks. $W_a$ and $W_s$ represent the parameters for aesthetic quality assessment and semantic recognition respectively. Each column in $W_a$ or $W_s$ corresponds to a subtask. The goal is to find the optimal or sub-optimal parameters $\Theta, W, \lambda$ by maximizing the following posterior probability
\begin{equation}
\hat{\Theta}, \hat{W}, \hat{\lambda} = \operatorname*{argmax}\limits_{\Theta, W, \lambda} p(\Theta, W, \lambda| X, Y, Z),
\end{equation}
where $\lambda$ is the weight coefficient of the semantic recognition task in the joint learning process.

Based on the Bayesian theorem, we have
\begin{equation}
\begin{split}
 p(\Theta, W, \lambda| X, Y, Z) &= \frac {p(X, Y, Z|\Theta, W, \lambda)p(\Theta, W, \lambda)}{p(X, Y, Z)} \\
 & \propto p(X, Y, Z|\Theta, W, \lambda)p(\Theta, W, \lambda),
 \end{split}
\end{equation}
where $p(X, Y, Z|\Theta, W, \lambda)$ is the conditional probability, and $p(\Theta, W, \lambda)$ is the prior probability.

Then Eqn.~(1) takes the form
\begin{equation}
\begin{split}
& \hat{\Theta}, \hat{W}, \hat{\lambda} \\
&  \propto \operatorname*{argmax}\limits_{\Theta, W, \lambda} p(Y|X, \Theta, W_a) p(Z|X, \Theta, W_s, \lambda) p(\Theta) p(W) p(\lambda).
\end{split}
\end{equation}
Each term in Eqn.~(3) is defined as:

1) The conditional probability $p(Y|X, \Theta, W_a)$ corresponds to the task of aesthetic quality assessment. Here assessing aesthetic quality is interpreted as a classification problem and modeled as a multinomial logistic regression similar to traditional classification problems~\cite{Krizhevsky12}. The conditional probability $p(Y|X, \Theta, W_a)$ can be formulated as
\begin{equation}
\begin{split}
p(Y|X, \Theta, W_a)=\prod_{n=1}^N \sum_{c=1}^C 1\{y_n=c\}p(y_n=c|x_n,\Theta, W_a),
 \end{split}
\end{equation}
where $1\{\cdot\}$ is the indicator function, it has two values, $1\{a\ true\ statement\}=1$, and $1\{a\ false\ statement\}=0$. $p(y_n=c|x_n,\Theta, W_a)$ is calculated by the softmax function
\begin{equation}
\begin{split}
p(y_n=c|x_n,\Theta, W_a)=\frac {\exp({W_a^c}^{\mathrm{T}} (\Theta ^{\mathrm{T}} x_n))}{\sum_{l=1}^C \exp({W_a^l}^{\mathrm{T}} (\Theta ^{\mathrm{T}} x_n ))}.
 \end{split}
\end{equation}

2) The conditional probability $p(Z|X, \Theta, W_s, \lambda)$ corresponds to the semantic recognition. Since each element of the semantic label of a given image is binary: $z_n^m \in \{0,1\}$, each semantic attribute recognition can be interpreted as a logistic regression. Hence the conditional probability $p(Z|X, \Theta, W_s, \lambda)$ can be
\begin{equation}
\begin{split}
&p(Z|X, \Theta, W_s, \lambda)\\
&=\prod_{n=1}^N \prod_{m=1}^M ( p(z_n^m=1|x_n,\Theta, W_s^m)^{z_n^m}\\
& (1-p(z_n^m=1|x_n,\Theta, W_s^m))^{1-z_n^m} )^{\lambda},
 \end{split}
\end{equation}
where $p(z_n^m=1|x_n,\Theta, W_s^m)$ is calculated by a sigmoid function $\sigma (x)=1/(1+\exp(-x))$.

3) The prior probability $p(\Theta)$ corresponds to the network parameters for common features. The parameters $\Theta$ can be initialized as a standard normal distribution like previous network~\cite{Krizhevsky12}. $p(\Theta)=\prod_{k=1}^K p(\theta_k)=\prod_{k=1}^K N(\textbf{0},I)$, where $\textbf{0}$ is a zero matrix and $I$ is an identity matrix.

4) Similar to $\Theta$, the parameters $W$ for specific tasks can also be initialized as a standard normal distribution. Thus, the prior probability can be $p(W)= p(W_a)p(W_s)= N_a(\textbf{0},I)N_s(\textbf{0},I)$.

5) $\lambda$ is used to control the influence of semantic recognition task in the final objective function. The prior probability $p(\lambda)$ is implemented by defining $\lambda$ obeying a normal distribution, $p(\lambda)=\emph{N}(\mu, \sigma^2)$.

Then Eqns.~(4),~(5) and~(6) are substituted into Eqn.~(3), negative log function is taken for Eqn.~(3), and the constant terms are omitted. As a result, the objective function can be
\begin{equation}
\begin{split}
 &\operatorname*{argmin}\limits_{\Theta, W, \lambda}  \{ - \sum_{n=1}^N \sum_{c=1}^C 1\{y_n=c\}log \frac {\exp({W_a^c}^{\mathrm{T}} (\Theta ^{\mathrm{T}} x_n))}{\sum_{l=1}^C \exp({W_a^l}^{\mathrm{T}} (\Theta ^{\mathrm{T}} x_n) )} \\
 & -\lambda \sum_{n=1}^N \sum_{m=1}^M ( z_n^m log \sigma({W_s^m}^{\mathrm{T}} (\Theta ^{\mathrm{T}} x_n))+ (1-z_n^m)(1-\\ &log\sigma({W_s^m}^{\mathrm{T}} (\Theta ^{\mathrm{T}} x_n))))+ \Theta ^{\mathrm{T}} \Theta + W ^{\mathrm{T}} W + (\lambda-\mu)^2\}.
 \end{split}
\end{equation}

\begin{figure*}
  \centering
  \includegraphics[width=16cm]{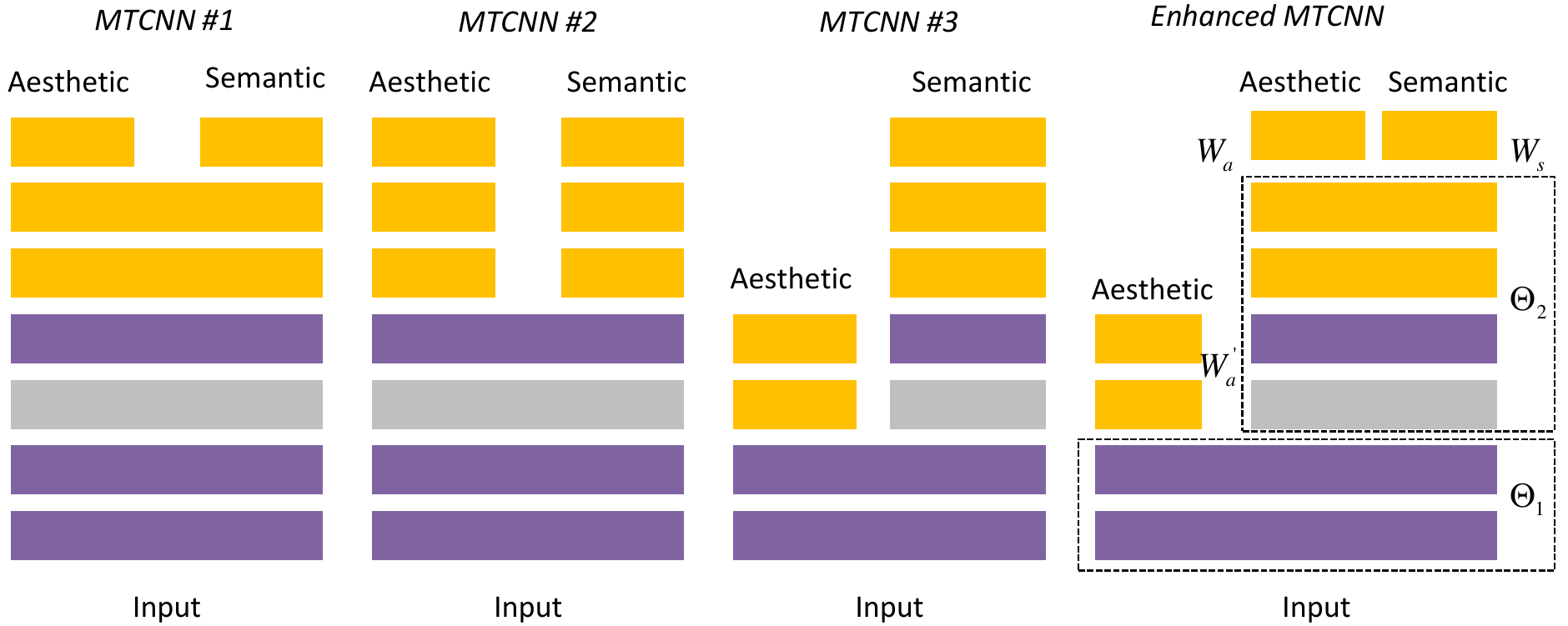}\\
  \caption{Explored MTCNNs with different architectures. The details of MTCNN \#1 are illustrated in Fig.~\ref{fig:MT1}. Color code used: purple = convolutional layer + max pooling, grey = convolutional layer, yellow = fully-connected layer.}
  \label{fig:mt4}
\end{figure*}
\subsection{Multi-Task Relationship Learning Probabilistic Framework \label{sec:MTRLPF}}
To automatically learn the relationships between aesthetic and semantic tasks and to better exploit the inter-task relatedness for aesthetic feature learning, we model the relationships between tasks as a covariance matrix $\Omega$ and add it to our above multi-task framework. The new framework is called Multi-Task Relationship Learning (MTRL) framework. In the MTRL framework, the goal is to find the optimal or sub-optimal parameters $\Theta, W, \lambda, \Omega$ by maximizing the following posterior probability
\begin{equation}
\hat{\Theta}, \hat{W}, \hat{\lambda} = \operatorname*{argmax}\limits_{\Theta, W, \lambda} p(\Theta, W, \lambda, \Omega| X, Y, Z),
\end{equation}
Based on the Bayesian theorem, Eqn.~(8) takes the form
\begin{equation}
\begin{split}
\hat{\Theta}, \hat{W}, \hat{\lambda} \propto \operatorname*{argmax}\limits_{\Theta, W, \lambda} &p(Y|X, \Theta, W_a) p(Z|X, \Theta, W_s, \lambda) \cdot \\
&p(W|\Omega) p(\Theta) p(W) p(\lambda).
\end{split}
\end{equation}
The conditional probability $p(Y|X, \Theta, W_a)$, the conditional probability $p(Z|X, \Theta, W_s, \lambda)$, the prior probability $p(\Theta)$, the prior probability $p(W)$ and the prior probability $p(\lambda)$ are same to the above definition in Section~\ref{sec:MTPF}. For the prior on the $W$, we consider two terms $p(W)$ and $p(W|\Omega)$. The prior probability $p(W)$ is to model the each column of $W$ as a standard normal distribution for each task and can separately penalize the complexity of the each column of $W$. The $p(W|\Omega)$ is to model the structure of $W$ between tasks by using a matrix-variate normal distribution~\cite{zhang2010convex,gupta1999matrix}. So we have
\begin{equation}
\begin{split}
p(W|\Omega) & =MN(0, I \otimes \Omega)\\
            & =\frac {exp(-\frac{1}{2} tr(I^{-1}W \Omega^{-1}W^T))}{(2 \pi)^{d(M+C)/2}|I|^{(M+C)/2}|\Omega|^{d/2}},
\end{split}
\end{equation}
where $d$ is the dimension of the common representation of all the tasks, such as the dimension of layer 7 in Fig.~\ref{fig:MT1}. The new objective function can be
\begin{equation}
\begin{split}
\operatorname*{argmin}\limits_{\Theta, W, \lambda} & \{ - \sum_{n=1}^N \sum_{c=1}^C 1\{y_n=c\}log \frac {\exp({W_a^c}^{\mathrm{T}} (\Theta ^{\mathrm{T}} x_n))}{\sum_{l=1}^C \exp({W_a^l}^{\mathrm{T}} (\Theta ^{\mathrm{T}} x_n) )} \\
 & -\lambda \sum_{n=1}^N \sum_{m=1}^M ( z_n^m log \sigma({W_s^m}^{\mathrm{T}} (\Theta ^{\mathrm{T}} x_n))+ (1-z_n^m)(1-\\ &log\sigma({W_s^m}^{\mathrm{T}} (\Theta ^{\mathrm{T}} x_n))))+ \Theta ^{\mathrm{T}} \Theta + W ^{\mathrm{T}} W+ (\lambda-\mu)^2 \\
 &+ tr(W \Omega^{-1}W^T) \}, \\
 s.t. \quad  & \Omega \geq 0, \quad   tr(\Omega)=1.
 \end{split}
\end{equation}
Where the constraint $tr(\Omega)=1$ is the same as in~\cite{zhang2010convex}.

\subsection{Optimization Procedure \label{sec:op}}
The multi-task objective function in Eqn.~(7) and (11) can be optimized by a network through stochastic gradient descent (SGD)~\cite{Krizhevsky12}. Here a specific CNN is applied to search optima for the parameters $\Theta, W, \lambda, \Omega$. One architecture of our MTCNNs is shown in Fig.~\ref{fig:MT1}. For the optimization procedure of MTCNNs, firstly, all tasks share knowledge in bottom layers. Then specific features are learned for each task in top layers. Finally, the combination of the softmax loss function for aesthetic quality prediction (the first term in Eqn.~(7)) and the cross entropy loss function for semantic recognition (the second term in Eqn.~(7)) are employed to update the parameters of the network jointly by back propagation. For the MTRLCNN, we adopt an alternate optimization procedure~\cite{zhang2010convex} to minimize the objective function in Eqn.~(11) for the parameters $\Theta, W, \Omega$. Firstly, we update $\Theta, W$ by back propagation like the MTCNN with fixed $\Omega$. Then fix $\Theta, W$ and optimize the $\Omega$, $\Omega=\frac{(W^TW)^{1/2}}{tr(W^TW)^{1/2})}$. We repeat this procedure until convergence.

Traditionally, multiple tasks are treated equally important in back propagation of multi-task learning~\cite{caruana1997,jung2015rotating} assuming that they can reach best performance roughly at the same time. However, different tasks may have different learning difficulties and convergence rates. Caruana~\cite{caruana1997} propose to control the effect of different tasks by adjusting the learning weight on each output task. He also put forward some strategies for this problem, such as early-stopping. Early stopping strategy has been used to some works~\cite{zhang2014facial} and good performance is achieved. Nevertheless, this strategy is not suited to our problem. This is because the extra task (i.e., semantic recognition task) is much easier, and often converges more quickly than the main task (i.e., aesthetic quality assessment). Our experimental results (details in Table~\ref{tab:table1} and Section~\ref{sec:exp}) show that, if the convergent semantic recognition task is early stopped, the training loss of the aesthetic task will do not drop obviously and converge in a low rate. We think that it is mainly because the aesthetic is subjective and needs the help of semantic task in entire training process. Hence, we present a simple strategy to keep the effect of all tasks balanced in back propagation. Because the softmax loss function only considers the value corresponding ground truth label for each example. In our problem, $\lambda=1/M$ is fixed in the objective function in the entire training process.

\subsection{Network Architectures Implementation Exploration}
To implement the multi-task model, we investigate several multi-task network architectures to utilize semantic information for visual aesthetic quality assessment. Take the MTCNN as an example and adapt the networks to our problem, then apply suited network architecture to our MTRLCNN. These networks are explained in Fig.~\ref{fig:mt4}. The supervision of aesthetic and semantic labels can be in the same or different layers in the network. Here we propose and explore three basic network architectures and an enhanced network. For all networks, the input is a $227 \times 227 \times 3$ patch randomly extracted from a resized image $256 \times 256 \times 3$ as previous work~\cite{lu14}.

\textbf{MTCNN \#1}: Since our goal is to discover the effective features for aesthetic assessment with the help of semantic information, a simple idea is to learn all parameters for aesthetic representations with aesthetic and semantic supervision in a network until the last layers. MTCNN \#1 implements this idea. The architecture of MTCNN \#1 (in Fig.~\ref{fig:mt4}) is detailed in Fig.~\ref{fig:MT1}. The network contains four convolutional layers and two fully-connected layers with parameters $\Theta$ for common feature learning. Then the network is split into two branches, the two last layers for two specific tasks. Thus the parameters $W=[W_a, W_s]$ from layer 6 to layer 7 for each task are learned separately. Then, the softmax loss function is adopted for aesthetic quality prediction, and the cross entropy loss function for semantic recognition. The combination of the two loss functions is employed to jointly update the parameters of the network.
\begin{figure*}
  \centering
  \includegraphics[width=16cm]{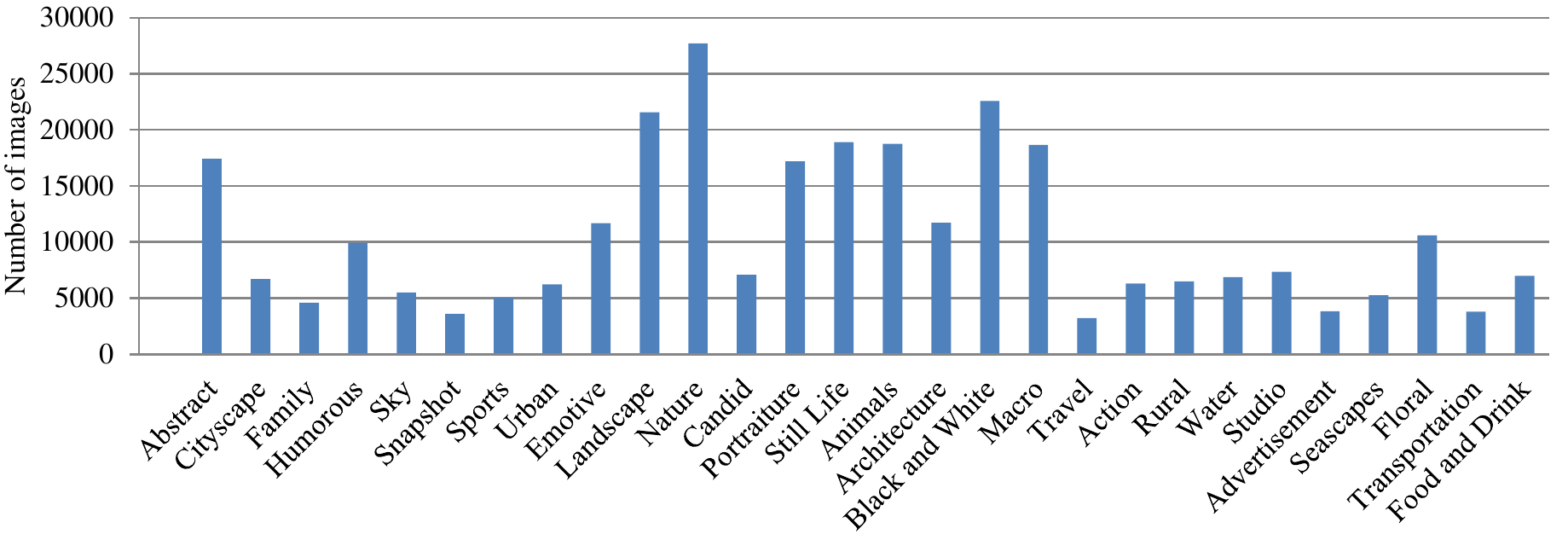}\\
  \vspace{-1em}
  \caption{The number of images for each semantic tag on AVA dataset.} \label{fig:NI}
\end{figure*}

\textbf{MTCNN \#2}: To explore different structures for aesthetic features learning, we introduce MTCNN \#2 (shown in Fig.~\ref{fig:mt4}) to allow some top layers to learn aesthetic representations independently without semantic supervision. Similar to MTCNN \#1, the network \#2 contains four convolutional layers with parameters $\Theta$ for common feature learning. Then the network is split into two branches earlier than MTCNN \#1 for two specific tasks. Different from the architecture \#1, layers 5, 6 and 7 in the network \#2 learn parameters $W=[W_a, W_s]$ separately for the two tasks. The loss functions are also the same as the architecture \#1.

\textbf{MTCNN \#3}: Since CNNs can learn hierarchical features, we consider the low-level features of a network for our main task in the MTCNN \#3 (shown in Fig.~\ref{fig:mt4}). In this network, four convolutional layers and three fully-connected layers are designed for semantic recognition, while two convolutional layers and two fully-connected layers for aesthetic quality assessment. The two tasks share knowledge $\Theta$ in the two convolutional layers. The other layers are used to learn specific parameters $W=[W_a, W_s]$ for each task. The loss functions are also the same as the architecture \#1.

\textbf{Enhanced MTCNN}: To further explore the effective aesthetic features, we propose an enhanced MTCNN by combining MTCNN \#1 and MTCNN \#3. That is, we add extra aesthetic supervision in the first two layers in MTCNN \#1. Shown in Fig.~\ref{fig:mt4}, the common parameters $\Theta_1$ in the first and second convolutional layers are learned for three tasks, the common parameters $\Theta_2$ in other two convolutional layers and two fully-connected layers are learned for two tasks, and specific parameters $W=[W_a^{'},W_a, W_s]$ are learned separately in top layers. Our goal is to enhance the supervision of aesthetic labels in the first and second convolutional layers under the premise of ensuring the influence of semantic information in all network. Here we denote $\Theta=[\Theta_1,\Theta_2]$. The objective function in Eqn.~(7) is transformed to
\begin{equation}
\begin{split}
 &\operatorname*{argmin}\limits_{\Theta, W, \lambda}  \{- \sum_{n=1}^N \sum_{c=1}^C 1\{y_n=c\}log \frac {\exp({W_a^c}^{\mathrm{T}} (\Theta ^{\mathrm{T}} x_n))}{\sum_{l=1}^C \exp({W_a^l}^{\mathrm{T}} (\Theta ^{\mathrm{T}} x_n) )} \\
 &- \sum_{n=1}^N \sum_{c=1}^C 1\{y_n=c\}log \frac {\exp({W_a^{'c}}^{\mathrm{T}} (\Theta_1 ^{\mathrm{T}} x_n))}{\sum_{l=1}^C \exp({W_a^{'l}}^{\mathrm{T}} (\Theta_1 ^{\mathrm{T}} x_n) )} \\
 & -\lambda \sum_{n=1}^N \sum_{m=1}^M ( z_n^m log \sigma({W_s^m}^{\mathrm{T}} (\Theta ^{\mathrm{T}} x_n))+ (1-z_n^m)(1-\\ &log\sigma({W_s^m}^{\mathrm{T}} (\Theta ^{\mathrm{T}} x_n))))+ \Theta ^{\mathrm{T}} \Theta + W ^{\mathrm{T}} W + (\lambda-\mu)^2\},\\
 \end{split}
\end{equation}
where the first term in Eqn.~(12) is our main task, and the second term is the added task. We fix $\lambda=2/M$ based on our strategy for the enhanced MTCNN.

\subsection{Transfer learning with semantic information}
Semantic content recognition has been studied for many years in computer vision, such as object recognition, object detection, image classification and semantic segmentation~\cite{Girshick_2015_ICCV,Krizhevsky12,long2015fully,Simonyan14c,He2015}. Recently, deep learning methods have achieved great succuss on the semantic recognition, especially the image classification on Imagenet~\cite{ Krizhevsky12,Simonyan14c,He2015,deng2009imagenet}. The Imagenet~\cite{deng2009imagenet} dataset contains rich semantic information and can be utilized to further help aesthetic representation learning. Thus we transfer the semantic representation learned from the network pretrained on Imagenet to aesthetic quality assessment. A trained model on a dataset can be transferred to another dataset for a similar or different task~\cite{pan2010survey,donahue2013decaf}. Specifically, our multi-task architecture from Layer 1 to Layer 6 in Fig.~\ref{fig:MT1} is replaced with AlexNet~\cite{Krizhevsky12}, VGG Net~\cite{Simonyan14c} or ResNet~\cite{He2015}. It is shown MTCNN \#1 performs best in the three basic MTCNNs from Table~\ref{tab:mt4res}. We initialize the networks with models pretrained on Imagenet and finetune it with the training data labeled with aesthetic labels and semantic labels.

In addition, another meaningful direction is how to exploit the massive dataset of visual semantic understanding for the limited dataset with only aesthetic labels for aesthetic assessment. To transfer the learned representation with both aesthetic and semantic supervision to the dataset with only aesthetic labels, we initialize the networks with pretrained multi-task models and finetune it with the training data labeled with only aesthetic labels.

\section{Experiments \label{sec:exp}}
In this section, we evaluate the proposed method on the challenging large-scale AVA dataset and Photo.net dataset. Experimental results show that the benefits of semantic information and the effectiveness of our proposed method.
\begin{figure*}
  \centering
  \includegraphics[width=18cm]{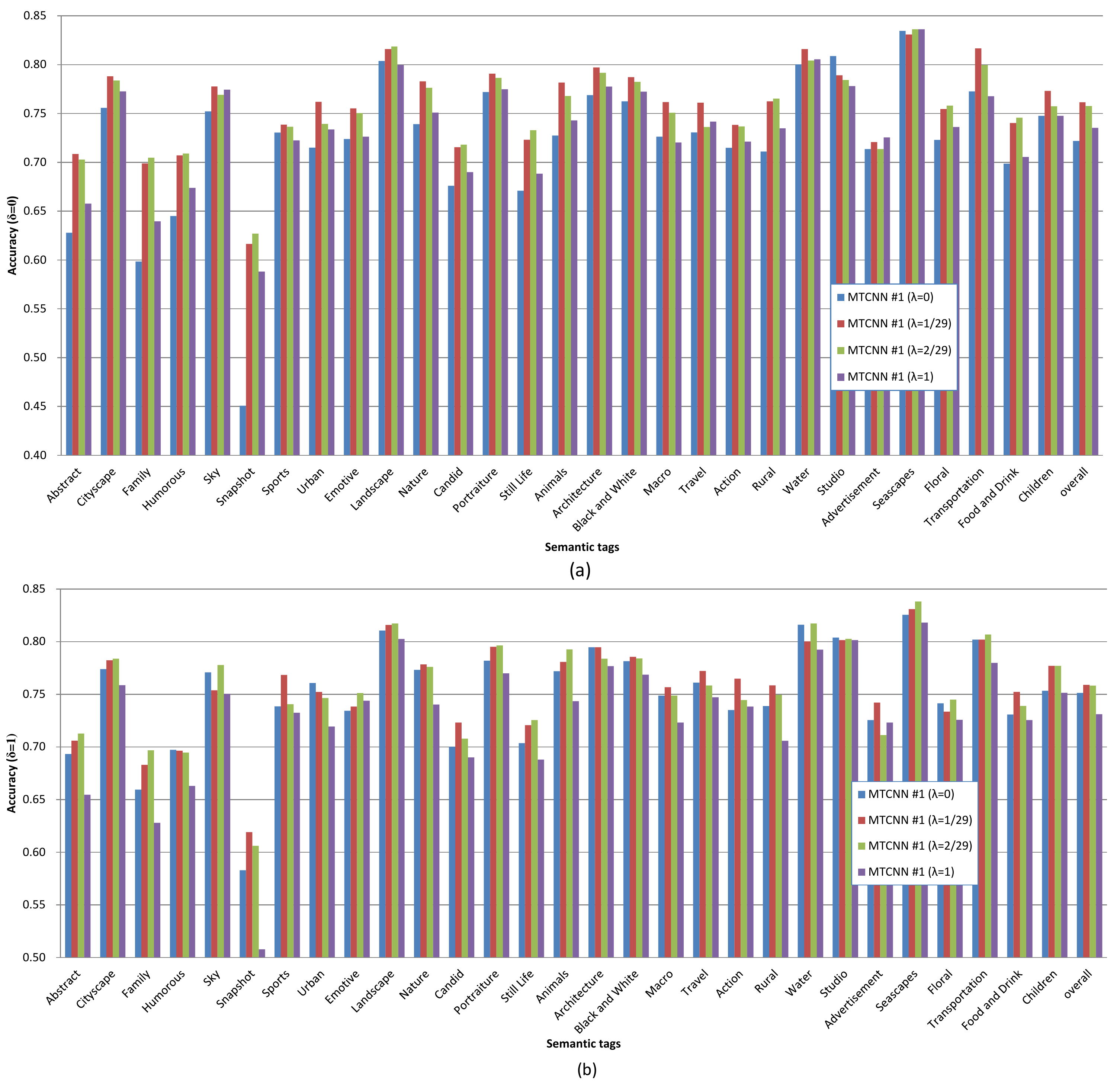}\\
  \caption{Accuracy on each semantic tag using MTCNN \#1 with different $\lambda$ when $\delta=0$ and $\delta=1$ on the AVA dataset.  }
  \label{fig:tag_r}
\end{figure*}
\subsection{Dataset}

\noindent\textbf{AVA dataset}: The AVA dataset~\cite{Murray12} is one of the most large-scale and challenging dataset for visual aesthetic quality assessment. It contains more than 255,000 images gathered from \emph{www.dpchallenge.com}. Each image has about 200 voters to assess the aesthetic score from one to ten. In addition, each image contains 0, 1 or 2 semantic tags (attributes). We select 185,751 images used in this paper based on the following rules. 1) More than 3000 images are available for each tag; 2) each image contains at least one tag. Eventually 29 semantic tags are chosen and the number of images for each tag is listed in Fig.~\ref{fig:NI}. From the 185,751 images, 20,000 images are randomly selected as the testing set similar to~\cite{lu14}, and the rest 165,751 images as the training set. For aesthetic labels, we follow the experimental setup as~\cite{Murray12,lu14}, the training set is divided into two classes: high quality and low quality images. We designate the images with an average score larger than $5 + \delta$ as high quality images, those with an average score smaller than $5 - \delta$ as low quality images. Images with an average score between $5 + \delta$ and $5 - \delta$ are discarded. We set $\delta$ to 0 and 1 respectively for the training set to obtain the ground truth labels. There are 165,751 images in the training set when $\delta=0$ and 38,994 images in the training set when $\delta=1$. We set $\delta$ to 0 for the testing set regardless of the value of $\delta$ for the training set. For semantic labels, each image is labeled as a 29-dim binary vector.

\begin{table}
\caption{Accuracy (\%) of our MTCNN \#1 with different $\lambda$  on  the AVA dataset.}
\renewcommand{\arraystretch}{1.4}
\setlength{\abovecaptionskip}{0pt}
\setlength{\belowcaptionskip}{0pt}
\arrayrulewidth=0.6pt \tabcolsep=4pt
  \centering
  \footnotesize
  \begin{tabular}{|c|c|c|c|c|c|c|c|}
    \hline
     $\delta$  & $\lambda=0$&$\lambda=1/29$&$\lambda=2/29$& $\lambda=1$& with early stopping\\
    \hline
    0 &72.19&\textbf{76.15}&75.76 &73.54& 73.43\\
     \hline
    1 &75.13&\textbf{75.90}&75.82&73.12& 74.28\\
    \hline
  \end{tabular}
  \verb''\\
  \label{tab:table1}
\end{table}
\begin{figure*}
  \centering
  \includegraphics[width=18cm]{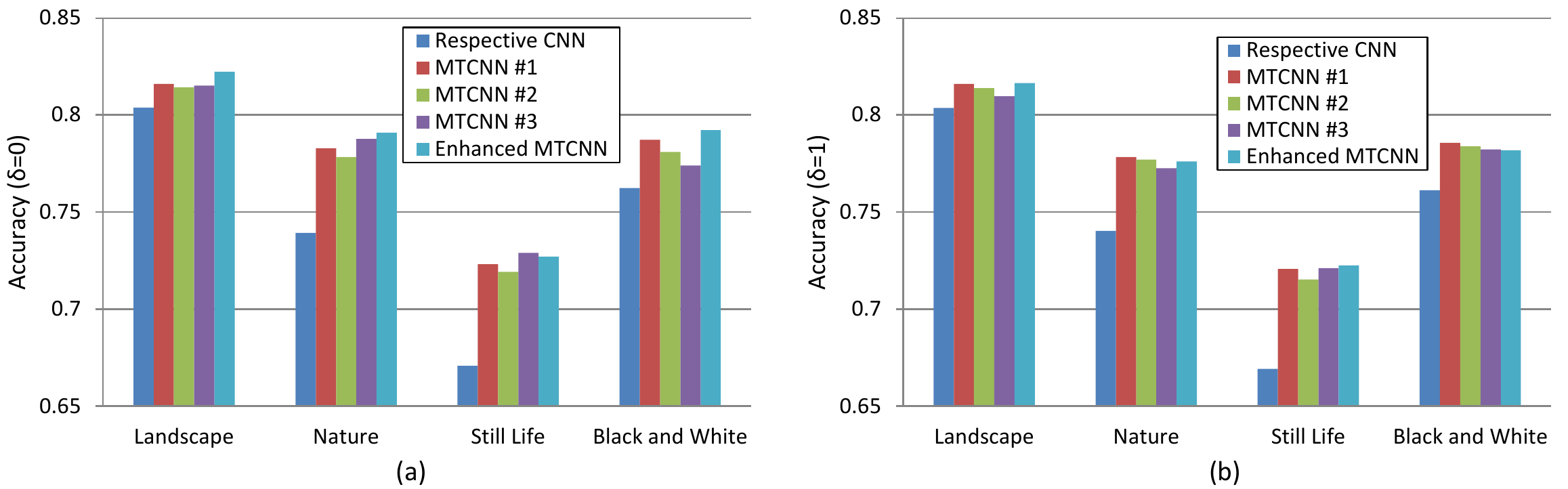}\\
  \caption{The accuracy with different methods for aesthetic classification on ``Landscape'', ``Nature'', ``Still Life'' and ''Black and White'' separately with both $\delta=0$ and $\delta=1$. }
  \label{fig:tag4}
\end{figure*}
\noindent\textbf{Photo.net dataset\footnote{Available at \emph{http://ritendra.weebly.com/aesthetics-datasets.html}}}: The Photo.net dataset~\cite{Datta08,Datta06} is a dataset with only aesthetic labels. It contains 20,278 images collected from \emph{www.photo.net}. Each image is rated by at least 10 users to assess the aesthetic quality from one to seven. Due to some missing images in the dataset, we collect 17,232 images in all. From the overall images, 3000 images are randomly selected as the testing set, and the rest 15,232 images as the training set. For the ground truth labels, we follow~\cite{Datta06} and choose the average score 5.0 as median aesthetic ratings. The images with an average score larger than $5 + \delta$ are designated as high quality images, those with an average score smaller than $5 - \delta$ as low quality images. We set $\delta$ to 0 in the experiment. Aesthetic quality assessment with $\delta=0$ is more challenging than that with $\delta>0$~\cite{Murray12}.

\subsection{Evaluating the Effectiveness of Keeping Balance Strategy}
In the objective function, $\lambda$ is used to control the contributions from semantic information. To validate our strategy of keeping the influence of two tasks balanced, we implement our MTCNN \#1 with our strategy $\lambda=1/M$ (here $\lambda=1/29$) and we also compare the experimental results of MTCNN \#1 with $\lambda=0$, $\lambda=2/29$, $\lambda=1$ and early stopping strategy (shown in Table~\ref{tab:table1}). By comparing the results with or without the supervision of semantic labels, the MTCNN \#1 with $\lambda \neq 0$ performs better than that with $\lambda=0$. This indicates the supervision is effective. What's more, the results shown in Table~\ref{tab:table1} demonstrate that our strategy $\lambda=1/29$ performs best on both values of $\delta$. When $\lambda=1/29$, the aesthetic and semantic tasks have same effect on the process of back propagation. Therefore the effectiveness of our strategy is verified.

To further demonstrate the effectiveness of our MTCNN with our strategy, we also analyze the accuracy on each semantic tag using MTCNN \#1 with different setting of $\lambda$ in Fig.~\ref{fig:tag_r}. As shown, our MTCNN \#1 with $\lambda=1/29$ performs best on overall images and most semantic tags. We also observe that different results are achieved on various semantic tags with the same method, and different improvements with MTCNNs are also different on various semantic tags. For example, the semantic tags ``Family'' and ``Snapshot'' obtain an great improvement with different methods.

 \begin{figure*}
  \centering
  \includegraphics[width=18cm]{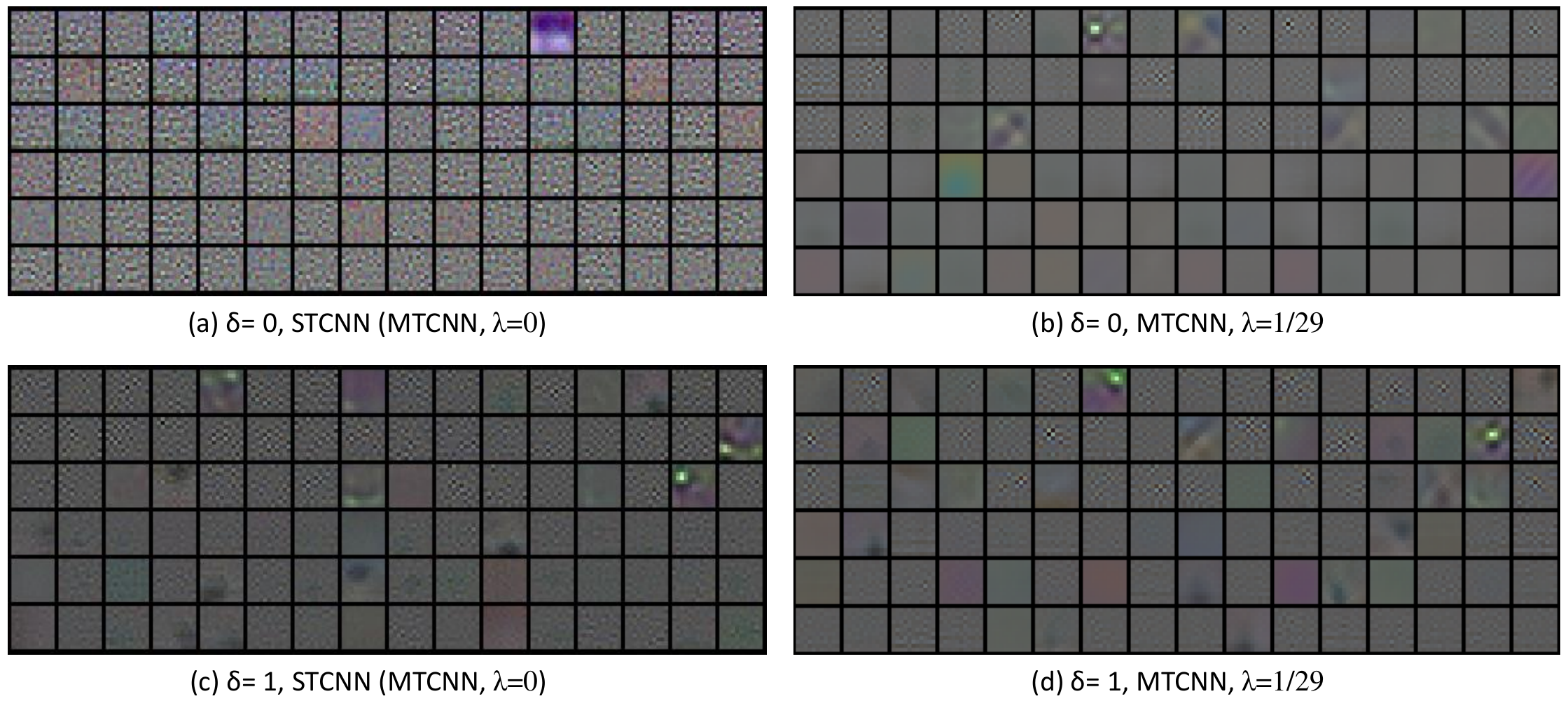}\\
  \caption{Learned filters in the first convolutional layer with STCNN for aesthetic task only and MCTNN \#1 for the two tasks with both $\delta=0$ and $\delta=1$. }
  \label{fig:filter}
\end{figure*}
\begin{figure*}
  \centering
  \includegraphics[width=18cm]{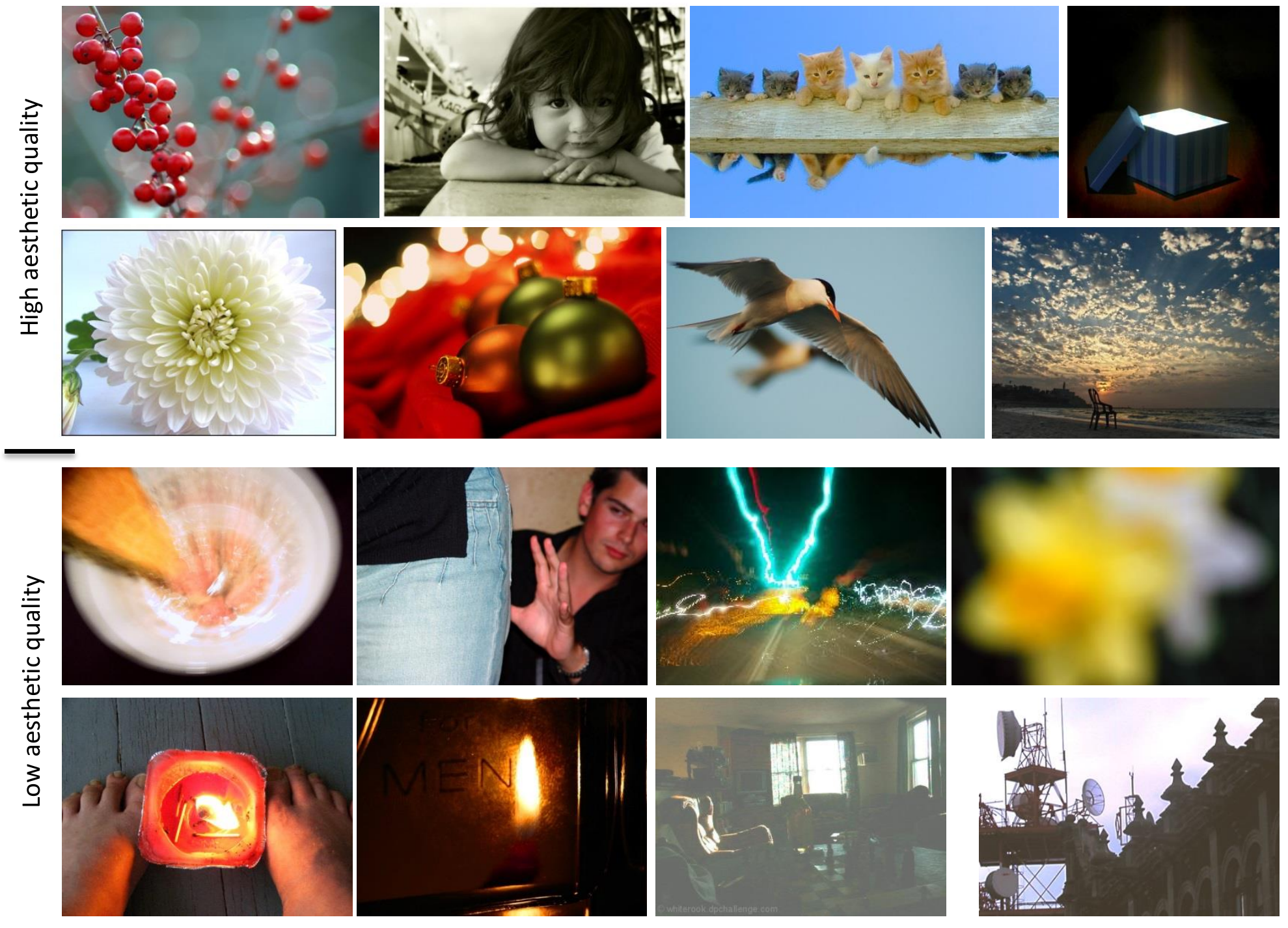}\\
  \caption{Example test images correctly classified by MTCNN but incorrectly by STCNN in the AVA dataset. The labels of the images on the first and second rows are high aesthetic quality, and the labels of the images on the third and fourth rows are low aesthetic quality. }
  \label{fig:examp}
\end{figure*}
\subsection{Evaluating the impact of network architectures}
To evaluate the impact of network architectures, we analyze the results with the three basic MTCNNs with $\lambda=1/M$ and enhanced MTCNN with $\lambda=2/M$ (shown in Table~\ref{tab:mt4res}). We can see that our enhanced MTCNN for the main task performs best. For the enhanced MTCNN, under the premise of ensuring the effect of semantic information in the whole network, we enhance the aesthetic supervision in the two bottom layers. Experimental results also show that MTCNN \#1 performs best in the three basic MTCNNs. Comparing the MTCNN \#1 and MTCNN \#2, we can see that late splitting obtains better performance for aesthetic quality assessment and semantic information is helpful for aesthetic representation learning. This also demonstrates that the more supervision semantic labels makes on the aesthetic feature learning, the better performance our MTCNN achieves. It also reveals that the low-level features of MTCNN \#3 can still perform well.

\begin{table}
\caption{Accuracy (\%) of four MTCNNs on the AVA dataset.}
\renewcommand{\arraystretch}{1.4}
\newcommand{\tabincell}[2]{\begin{tabular}{@{}#1@{}}#2\end{tabular}}
\setlength{\abovecaptionskip}{0pt}
\setlength{\belowcaptionskip}{0pt}
\arrayrulewidth=0.6pt \tabcolsep=3pt
  \centering
  \small
  \begin{tabular}{|c|c|c|c|c|c|c|c|c||c|c|c|c|c|c}
    \hline
     $\delta$  & MTCNN  \#1 & MTCNN \#2 & MTCNN \#3 & Enhanced MTCNN \\
    \hline
    0 &76.15&75.91 & 75.92& 76.58 \\
     \hline
    1 &75.90&75.81& 75.37&76.04\\
    \hline
  \end{tabular}
  \verb''\\
  \label{tab:mt4res}
\end{table}

\subsection{Evaluating the Benefits of Semantic Information }
To evaluate our MTCNNs with the help of semantic information for aesthetic classification, we compare our results of four MTCNNs with those of our single task CNN (STCNN, MTCNN \#1, $\lambda=0$) on the AVA dataset with both values of $\delta$. Shown in Table~\ref{tab:mt4res} and Table~\ref{tab:all}), all the four MTCNNs perform better than our STCNN especially when $\delta=0$. Aesthetic quality classification with $\delta=0$ is more challenging than that with $\delta=1$~\cite{Murray12}. These results demonstrate the effectiveness of semantic information.

Furthermore, we also train a separate model for each semantic labels to assess aesthetic quality. Due to different number of images for different semantic labels, we only train four CNNs separately for ``Landscape'', ``Nature'', ``Still Life'' and ''Black and White''. The four labels have the most number of images in 29 labels. Here we call the CNNs trained separately for the four semantic labels ``respective CNN''. For example, the respective CNN for ``Landscape'' is trained only with ``Landscape'' images for aesthetic categorization. Figure~\ref{fig:tag4} shows the results with different methods for aesthetic classification on ``Landscape'', ``Nature'', ``Still Life'' and ''Black and White'' separately with both value of $\delta$. As shown in Fig.~\ref{fig:tag4}, all the MTCNNs outperform the respective CNN on each semantic labels, which also demonstrates the effectiveness of semantic information for representation learning. Moreover, MTCNNs don't need to know the semantic labels of the testing images, while the respective CNNs have to know the semantic labels.

\begin{figure}
  \centering
  \includegraphics[width=9cm]{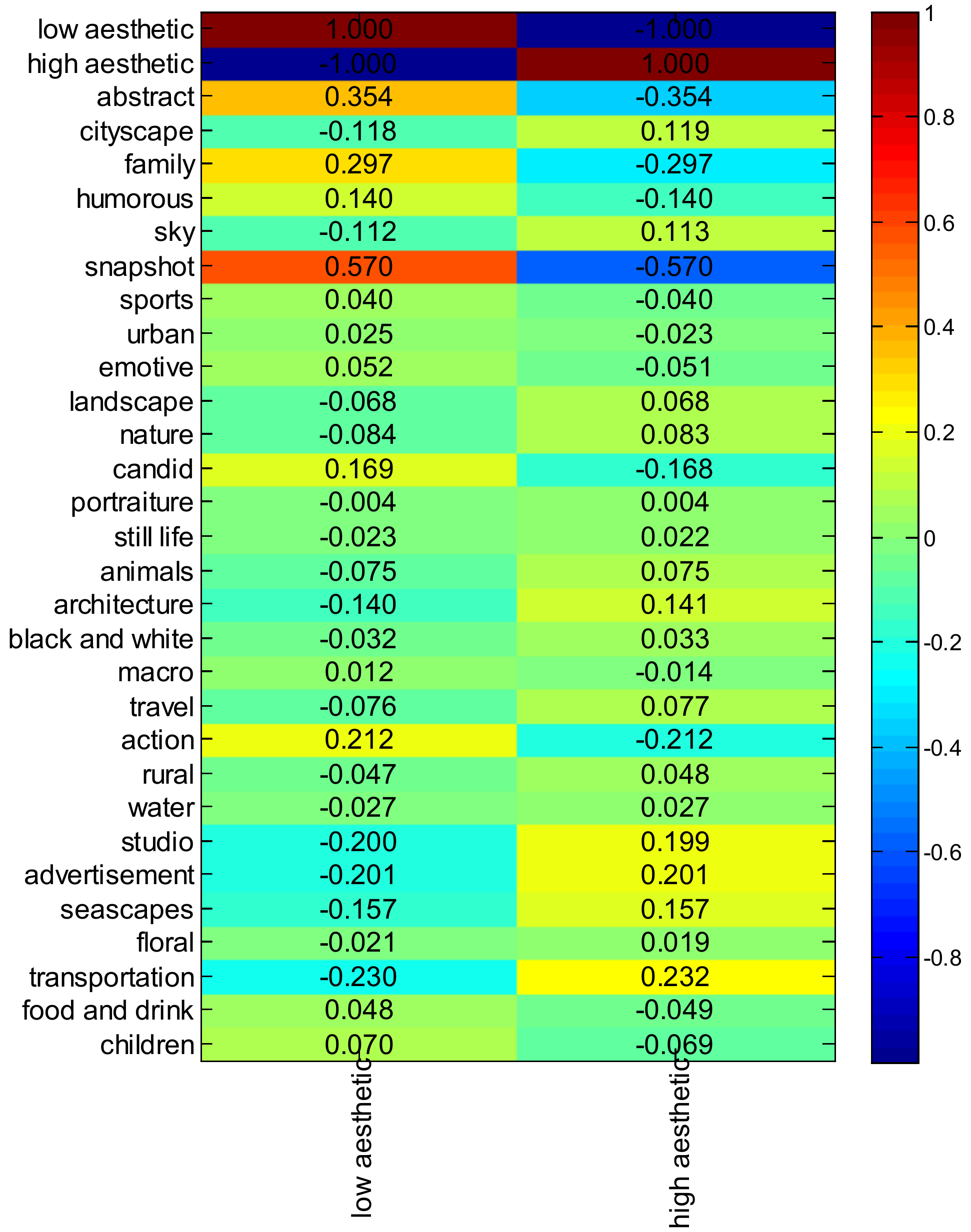} \\
  \caption{Correlation in any two subtasks of aesthetic quality classification and semantic recognition learned by MTRLCNN \#1 with $\delta=0$.}
  \label{fig:corr}
\end{figure}
To qualitatively demonstrate the benefits of our MTCNN with semantic information, we show learned filters in the first convolutional layer with a STCNN for aesthetic task only and our MCTNN \#1 with both $\delta=0$ and $\delta=1$ in Fig.~\ref{fig:filter}. Compared to the filters learned without semantic information, the filters with semantic information are smoother, cleaner and more understandable. The proposed MTCNN can learn more color and high frequency edge information than STCNN. These differences can also be observed from the examples of test images correctly classified by MTCNN but misclassified by STCNN in Fig.~\ref{fig:examp}. The high quality images often have more vivid color and clearer edge than low quality images. Most of the low quality images in Fig.~\ref{fig:examp} are blurred and dull. This indicates that the supervision of semantic labels for aesthetic feature learning is very beneficial, and aesthetic and semantic tasks are related to some extent.

To exploit the semantic information in the Imagenet, we select the late splitting multi-task network (such as MTCNN \#1) and replace the MTCNN \#1 architecture from Layer 1 to Layer 6 in Fig.~\ref{fig:MT1} with AlexNet~\cite{ Krizhevsky12}, VGG Net~\cite{Simonyan14c} or ResNet~\cite{He2015} respectively. That is because that the MTCNN \#1 performs best in the three basic MTCNNs. The networks are initialized with models pretrained on Imagenet and finetuned with the training data labeled with aesthetic labels and semantic labels. Table~\ref{tab:relation} shows the results of the three MTCNN networks (AlexNet\_FT, VGG Net\_FT and ResNet\_FT) with finetuning. It demonstrates the effectiveness of semantic information in Imagenet dataset. By comparing among three pre-trained networks, especially the ResNet~\cite{He2015}, the deeper network learns more semantic representation and performs better for aesthetic quality assessment by transfer learning.

\begin{table}
\caption{Accuracy (\%) of different network with or without relationship learning on the AVA dataset.}
\renewcommand{\arraystretch}{1.4}
\newcommand{\tabincell}[2]{\begin{tabular}{@{}#1@{}}#2\end{tabular}}
\setlength{\abovecaptionskip}{0pt}
\setlength{\belowcaptionskip}{0pt}
\arrayrulewidth=0.6pt \tabcolsep=2pt
  \centering
  \small
  \begin{tabular}{|c|c|c|c|c|c|c|c|c|c|c|c}
    \hline
     $Architecture $ & MTCNN \#1  &AlexNet\_FT & VGG Net\_FT & ResNet\_FT \\
    \hline
    MTCNN     &76.15     & 76.70 &77.73 & \textbf{78.56 }  \\
     \hline
    MTRLCNN   &76.56     & 77.35 &78.46 & \textbf{79.08 }  \\
    \hline
  \end{tabular}
  \verb''\\
  \label{tab:relation}
\end{table}
\begin{table*}
\caption{Accuracy (\%) of different methods on the AVA dataset.}
\renewcommand{\arraystretch}{1.4}
\newcommand{\tabincell}[2]{\begin{tabular}{@{}#1@{}}#2\end{tabular}}
\setlength{\abovecaptionskip}{0pt}
\setlength{\belowcaptionskip}{0pt}
\arrayrulewidth=0.6pt \tabcolsep=3pt
\centering
\small
\begin{tabular}{|c|c|c|c|c|c|c|c|c|c|c|c}
\hline
 $\delta$ & \tabincell{c}{Our\\ STCNN} & MTCNN \#1 & \tabincell{c}{MTRLCNN \\AlexNet\_FT} & \tabincell{c}{MTRLCNN \\VGG Net\_FT} & \tabincell{c}{MTRLCNN \\ResNet\_FT} & ~\cite{Murray12} & SCNN~\cite{lu14}& RDCNN~\cite{lu14}& DMA-Net~\cite{lu2015deep} &\tabincell{c}{MNA-CNN~\cite{mai2016composition} \\(VGG Net\_FT)} \\
\hline
0 &72.19&76.15     & 77.35&78.46 & \textbf{79.08 }   &66.7  & 71.20&74.46&75.41&77.4\\
 \hline
1 &75.13&75.90     & 76.80&77.41 & \textbf{77.71}    &67.0  & 68.63&73.70 &--   &76.5\\
\hline
\end{tabular}
\verb''\\
\label{tab:all}
\end{table*}
\subsection{Inter Tasks Correlation Analysis}
To further demonstrate the effectiveness of semantic information and investigate how semantic information influence aesthetic task again, we analyze the correlation between the two tasks. Since each column vector of task-specific matrix $W=[W_a, W_s]$ in the network corresponds to the parameters of a subtask, we use the learned covariance matrix $\Omega$ and calculate the correlation coefficient between any two subtasks~\cite{fan2002sas}. Shown in layer 7 of Fig.~\ref{fig:MT1} in our problem, the aesthetic classification task has two subtasks: high aesthetic and low aesthetic, the semantic recognition task has 29 subtasks. Figure~\ref{fig:corr} presents the correlation between the aesthetic subtasks and sematnic subtasks learned by MTRLCNN \#1 with $\delta=0$, which also verifies that semantic information is beneficial for aesthetic estimation. Seen from Fig.~\ref{fig:corr}, a low aesthetic task has high negative correlation with a high aesthetic task. We can also see that the aesthetic tasks have high correlation with certain semantic attributes. For instance, the semantic tags ``Snapshot'' and ``Candid'' recognition has high positive correlation with the low aesthetic task. In real word, most of ``Snapshot'' and ``Candid'' images are usually regarded as low aesthetic quality images. While ``Advertisement'' and ``Seascapes'' recognition has positive correlation with the high aesthetic task. This accords with the knowledge that most of ``Seascapes'' and ``Advertisement'' images are usually taken as high aesthetic quality images. In addition, Fig.~\ref{fig:corr} can also visualize the correlation in different semantic tag recognitions. We also present the results of networks with or without relationship learning for aesthetic quality assessment in Table~\ref{tab:relation}, which validates the task relationship learning.

\subsection{Comparison with Other State-of-the-art Methods  }
To further validate our method with semantic information for aesthetic classification, we compare our results with those of the state-of-the-art methods in~\cite{Murray12,lu14,lu2015deep,mai2016composition} on the AVA dataset. Shown in Table~\ref{tab:mt4res} and Table~\ref{tab:all}, all the multi-task models perform better than the method in~\cite{Murray12}, SCNN~\cite{lu14}, and RDCNN~\cite{lu14} in on both values of $\delta$. The method in~\cite{Murray12} is the baseline of the AVA dataset and is implemented by extracting fisher vector (FV) descriptors~\cite{FV07} on the top of SIFT~\cite{Mar11} information and SVM classifier~\cite{CC01a}. SCNN is a single-column CNN, and RDCNN is a double-column CNN with an aesthetic column and a pretrained style column. Our results of MTRLCNN with VGG net and ResNet finetuning outperform the state-of-the-art method~\cite{mai2016composition}. Thus, these results in Table~\ref{tab:mt4res} and Table~\ref{tab:all} illustrate the effectiveness of our method with semantic recognition task.

Since the name list of 20,000 testing images used in~\cite{Murray12,lu14,lu2015deep,mai2016composition} are unavailable, the 20,000 images for testing in this paper maybe potentially different from the 20,000 testing images in~\cite{Murray12,lu14,lu2015deep,mai2016composition}. Thus, we performed 4 times with similar operation (20,000 images are randomly selected for testing at each time) for MTCNN \#1 ($\lambda=1/29, \delta=0$). The mean and variance (76.25\%, 0.0066) are close to our 76.15\%, which shows the robustness of our method. In addition, in this paper we selects 185,751 training images according to some rules, including the rule that all images need to have at least one semantic tag. It seems that the our training set is more clean than the 230,000 training images in~\cite{Murray12,lu14,lu2015deep,mai2016composition} and maybe helpful. To clarify how much benefit our method training with a ``clean'' set, we implement the baseline model (STCNN) trained on the full training set of 230,000 images. The accuracies on the same test set are 72.20\% ($\delta=0$), 75.27\% ($\delta=1$) and close to 72.19\% ($\delta=0$), 75.15\% ($\delta=1$) with a ``clean'' set. It seems that training with a ``clean'' set does not help the current method. This also demonstrates that our multi-task models with smaller training data can still outperform the state-of-the-art methods.

Although our goal is to improve the performance of aesthetic quality assessment without considering the evaluation of semantic task, we also give the 64.89\% Average Precision of MTCNN\#1 ($\lambda=1/29,\delta=0$) and 67.44\% of MTRLCNN with ResNet\_FT ($\lambda=1/29,\delta=0$).
\begin{table}
\caption{Accuracy (\%) of different methods on the Photo.net dataset.}
\renewcommand{\arraystretch}{1.4}
\newcommand{\tabincell}[2]{\begin{tabular}{@{}#1@{}}#2\end{tabular}}
\setlength{\abovecaptionskip}{0pt}
\setlength{\belowcaptionskip}{0pt}
\arrayrulewidth=0.6pt \tabcolsep=1pt
  \centering
  \small
  \begin{tabular}{|c|c|c|c|c|c|c|c|c|c|}
    \hline
     $\delta$&GIST\_SVM& FV\_SIFT\_SVM & STCNN& STCNN\_FT& MTCNN \#1\_FT   \\
    \hline
    0 &59.90& 60.80 &61.00 & 62.10& \textbf{65.20}\\
     \hline
  \end{tabular}
  \verb''\\
  \label{tab:photo}
\end{table}
%

\subsection{Evaluating the Transfer Learning for Photo.net Dataset}
\begin{figure*}
  \centering
  \includegraphics[width=18cm]{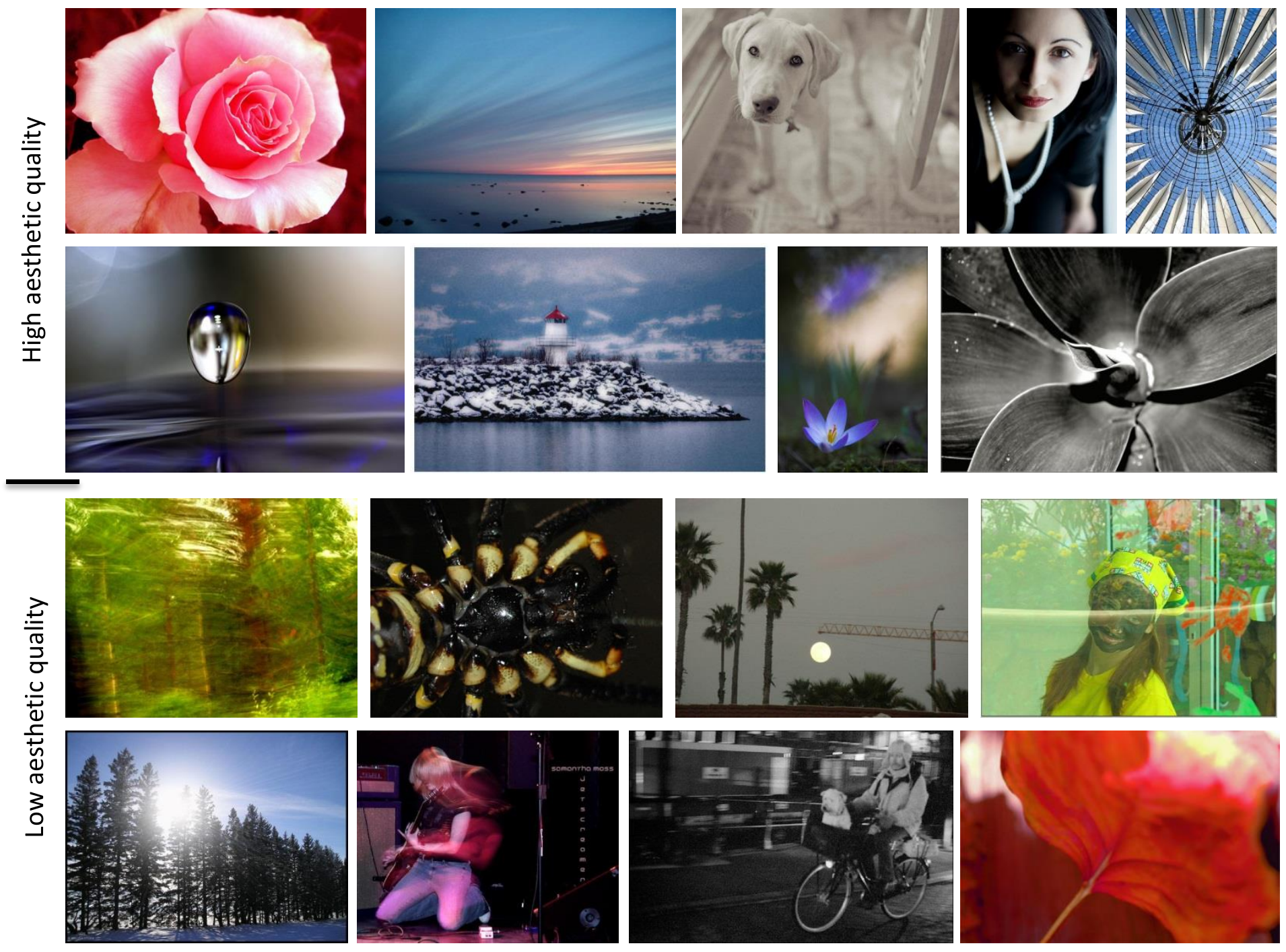}\\
  \caption{Example test images correctly classified by MTCNN \#1\_FT but incorrectly by STCNN\_FT in the Photo.net dataset. The labels of the images on the first and second rows are high aesthetic quality, and the labels of the images on the third and fourth rows are low aesthetic quality. }
  \label{fig:emp_photo}
\end{figure*}

 To utilize the semantic information for the dataset with only aesthetic labels, we transfer the learned representation with both aesthetic labels and semantic labels for the dataset with only aesthetic labels. In this paper, we exploit the learned representation with aesthetic and semantic labels from AVA dataset in MTCNN \#1 and finetune it with Photo.net dataset with only aesthetic labels. We call this model as MTCNN \#1\_FT. To validate the effectiveness of transferred representation with semantic information, we finetune the pretrained STCNN model on AVA dataset with only aesthetic labels for Photo.net dataset (STCNN\_FT). Moreover, we also train a STCNN on Photo.net dataset without finetuning. Furthermore, we implement the GIST descriptors~\cite{Oliva01} and FV on the top of SIFT with a SVM classifier (GIST\_SVM and FV\_SIFT\_SVM). Table~\ref{tab:photo} shows the accuracy of these methods on Photo.net dataset. Fig.~\ref{fig:emp_photo} visualizes some testing images correctly classified by MTCNN \#1\_FT but incorrectly by STCNN\_FT in the Photo.net dataset. These reveal the effectiveness of transfer learning with semantic information.

\section{Conclusion and Future Work \label{sec:con}}
In this paper, we have employed the semantic information to help discover representations for aesthetic quality assessment by formulating an end-to-end multi-task deep learning framework. Aesthetic quality assessment has not been taken as an isolation problem. To make full use of the semantic information and investigate how semantic information influence aesthetic task, four MTCNNs have been explored to learn the aesthetic representation jointly with the supervision of aesthetic and semantic labels. At the same time, a strategy of keeping the effect of two tasks balanced is presented to optimize the parameters of our multi-task networks. In addition, task relationship learning is modeled in the multi-task framework and the correlations in the two tasks have been learned to investigate the role of semantic recognition in aesthetic quality assessment. Experimental results have shown that our method performs better than the state-of-the-art methods. It is demonstrated that the semantic information is beneficial to aesthetic feature learning and the high-level features in the network play an important role in aesthetic quality assessment. 

Although the proposed multi-task framework results in state-of-the-art results on the challenging dataset, how to perform aesthetic quality assessment like a human brain is still an ongoing issue. Future work is to explore other possible solutions to efficiently utilize the aesthetic and semantic information in a brain-like way. Another possible trend is to discover more possible and potential factors to affect aesthetic quality assessment.

%
%

\ifCLASSOPTIONcaptionsoff
  \newpage
\fi



\bibliographystyle{IEEEtran}
\bibliography{mybibfile}
\end{document}